\documentclass[10pt, conference, compsocconf]{IEEEtran}
\ifCLASSINFOpdf
   \usepackage[pdftex]{graphicx}
\else
\fi
\usepackage{url}

\begin{document}
%
\title{STAR-Net: Action Recognition using Spatio-Temporal Activation Reprojection}


\author{\IEEEauthorblockN{William McNally, Alexander Wong, John McPhee}
\IEEEauthorblockA{Dept. of Systems Design Engineering\\
University of Waterloo\\
Waterloo, Canada, N2L 3G1\\
\{wmcnally, a28wong, mcphee\}@uwaterloo.ca}
}


%


\maketitle

\begin{abstract}
While depth cameras and inertial sensors have been frequently leveraged for human action recognition, these sensing modalities are impractical in many scenarios where cost or environmental constraints prohibit their use.  As such, there has been recent interest on human action recognition using low-cost, readily-available RGB cameras via deep convolutional neural networks. However, many of the deep convolutional neural networks proposed for action recognition thus far have relied heavily on learning global appearance cues directly from imaging data, resulting in highly complex network architectures that are computationally expensive and difficult to train. Motivated to reduce network complexity and achieve higher performance, we introduce the concept of spatio-temporal activation reprojection (STAR). More specifically, we reproject the spatio-temporal activations generated by human pose estimation layers in space and time using a stack of 3D convolutions.  Experimental results on UTD-MHAD and J-HMDB demonstrate that an end-to-end architecture based on the proposed STAR framework (which we nickname STAR-Net) is proficient in single-environment and small-scale applications. On UTD-MHAD, STAR-Net outperforms several methods using richer data modalities such as depth and inertial sensors.

\end{abstract}


%
\IEEEpeerreviewmaketitle

\section{Introduction}
Human action recognition has been a popular research focus for several decades due to its wide range of applications in intelligent video surveillance \cite{ji20133d}, sports analytics \cite{fani2017hockey}, and human-computer interaction \cite{mitra2007gesture}. With recent innovations in sensor technology, new data modalities such as 3D skeletal coordinates obtained from depth cameras, and wearable inertial sensors, have been explored for the purpose of enhancing action recognition performance \cite{chen2016fusion, du2015hierarchical, liu2016spatio}. However, the practical limitations of these sensors render many applications infeasible. Depth cameras are severely limited by their working range, often fail in outdoor scenes due to sunlight interference \cite{mehta2017vnect}, and are not as widely available or economically viable as RGB cameras. Wearable inertial sensors must be worn by the subject, prohibiting ``in-the-wild'' applications like sports analytics and intelligent surveillance. For these reasons, performing action recognition strictly using RGB images remains highly desirable. 

\begin{figure}
    \centering
    \includegraphics[width=0.9\linewidth]{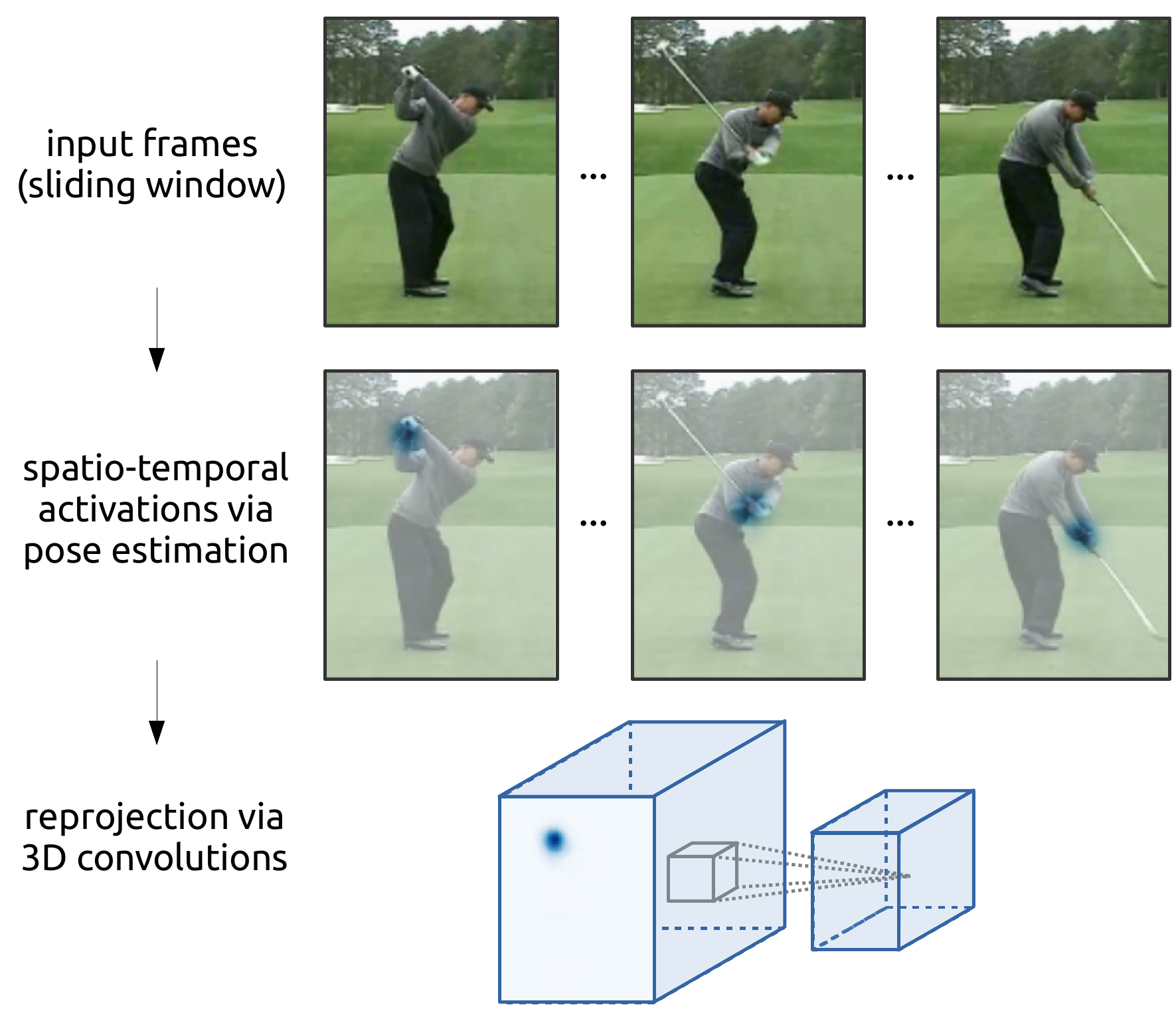}
    \vspace{-10pt}
    \caption{The STAR framework. Input frames are extracted from a video sample using a sliding window. Spatio-temporal activations representing the motions of joints (left wrist shown in the figure) are generated using a stack of pose estimation layers. The activations are then reprojected in space and time using a stack of 3D convolutions.}
    \label{fig:teaser}
    \vspace{-15pt}
\end{figure}

Recently, the ubiquity of RGB video data from online sources has fostered large-scale action recognition datasets that have spurred the use of deep learning and eliminated the need for engineered features. With the success of deep convolutional neural networks (CNNs) for visual recognition tasks~\cite{krizhevsky2012imagenet}, RGB-based action recognition has shifted towards using deep CNNs to data-mine spatio-temporal features in image sequences. Incorporating temporal information into CNNs has been accomplished using 3D convolutions~\cite{ji20133d}, recurrent networks \cite{donahue2015long, peng2016multi}, or by fusing spatial and temporal features from multiple streams (\textit{i.e.}, RGB and optical flow)~\cite{karpathy2014large}. Still, these action recognition models have limitations. First, they rely heavily on global appearance cues and could potentially underperform in situations where multiple unique actions exist within a single environment (\textit{e.g.}, in a sports match). Second, they are highly complex and contain a large number of parameters. As a result, training requires large amounts of data and is computationally taxing.

In a parallel line of computer vision research, CNNs have been used extensively to infer 2D human pose from RGB images \cite{newell2016stacked, cao2017realtime, chen2017cascaded}. Although these two streams of research share many similarities, utilizing the \textit{spatial activations} generated within human pose estimation networks for action recognition remains largely unexplored. 

Motivated by this, we introduce the concept of spatio-temporal activation reprojection (STAR) for action recognition.  More specifically,  spatio-temporal activations generated by a stack of pose estimation layers are reprojected in space and time using a stack of 3D convolutions that have been learned directly from the data (see Fig.~\ref{fig:teaser}). By leveraging spatio-temporal activations that are linked to human pose, the STAR framework is not influenced by global appearance cues, making it better suited for single-environment applications where variation in human movement is critical. Moreover, initializing the network with pretrained pose estimation layers shortcuts a large portion of complex spatial learning, permitting a network built around the STAR framework to be trained quickly and with limited data.  We empirically demonstrate that superior performance under said conditions can be achieved using an end-to-end network architecture based on the STAR framework (which we call STAR-Net) through evaluation on UTD-MHAD~\cite{chen2015utd} and J-HMDB~\cite{jhuang2013towards}, two small-scale action recognition datasets. On UTD-MHAD, STAR-Net outperforms several methods using richer data modalities, including methods leveraging depth images, inertial sensor data, and 3D skeletal coordinates. 

\section{Related Work}

\subsection{Action Recognition using RGB Images} 

Before the prevalence of CNNs for computer vision tasks, action recognition was performed by extracting features from RGB image sequences using traditional image processing techniques. Blank \textit{et al.}~\cite{blank2005actions} regarded human actions as 3D shapes by extracting silhouettes from each frame and forming space-time volumes. Sch\"uldt \textit{et al.}~\cite{schuldt2004recognizing} integrated local space-time features with support vector machines. Many generic image descriptors have been extended to video for the purpose of action recognition, including 3D-SIFT~\cite{scovanner20073}, HOG3D~\cite{klaser2008spatio}, extended SURF~\cite{willems2008efficient}, and Local Trinary Patterns~\cite{yeffet2009local}. Optical flow techniques employing dense feature trajectories have also demonstrated success for action recognition applications \cite{wang2013action}.

Recently, CNNs have been shown to be extremely proficient at image recognition~\cite{krizhevsky2012imagenet}. Successful CNN architectures have been adapted to accommodate video using 3D convolutions~\cite{ji20133d}, long-short term memory (LSTM)~\cite{donahue2015long}, and two-stream approaches using RGB and optical flow~\cite{karpathy2014large}. 

Three-dimensional CNNs (3D CNNs) extract features from both the spatial and temporal dimensions by convolving 3D filters across temporally-stacked image sequences, thereby capturing motion information across adjacent frames. Compared to LSTM and two-stream models, 3D CNNs are attractive because they directly create hierarchical representations of spatio-temporal data. Ji \textit{et al.}~\cite{ji20133d} used 3D CNNs for intelligent surveillance and achieved superior performance in comparison to baseline methods. Tran \textit{et al.}~\cite{tran2015learning} used 3D CNNs for large-scale video classification and found that small 3x3x3 filters were most effective. Carreira and Zisserman~\cite{carreira2017quo} demonstrated how 2D CNNs could be ``inflated'' to 3D, making it possible to extract spatio-temporal features from video while leveraging proven 2D CNN architectures and even their parameters. However, 3D CNN architectures cannot be warm-started on ImageNet. Moreover, 3D CNNs contain many more parameters than their 2D counterpart due to the extra filter dimension. For these reasons, monolithic 3D CNNs are prone to overfitting and require enormous amounts of video data to achieve good results~\cite{carreira2017quo}. Training difficulties are mitigated in this work using a shallow 3D CNN that takes refined spatio-temporal information as input.

\subsection{Action Recognition using Pose Data} 

Three-dimensional skeletal coordinates obtained from depth cameras have frequently been leveraged to perform action recognition using both hand-crafted and machine-learned features. Examples of hand-crafted features include the relative positioning between joints~\cite{wang2012mining}, covariance matrices of joint trajectories \cite{hussein2013human}, and view-invariant histograms~\cite{xia2012view}. With the increasing popularity of deep learning approaches, recurrent neural networks employing LSTM blocks have been used extensively to encode temporal sequences of 3D skeletal data~\cite{du2015hierarchical, liu2016spatio}. In other works, CNNs were used to recognize joint trajectory maps~\cite{wang2016action, hou2018skeleton}. The major drawback of these methods is that they rely on 3D skeletal coordinates obtained from the Microsoft Kinect, rendering them impractical in many scenarios. 

There are but a few very recent works that perform human action recognition using pose information extracted from RGB images. Yan \textit{et al.}~\cite{yan2018spatial} used a 2D human pose estimation model~\cite{cao2017realtime} as a standalone toolbox to extract joint coordinates and use them as input to a graph convolutional network. Liu \textit{et al.}~\cite{liu2018recognizing} used a standalone 2D human pose estimation model to recognize actions as the evolution of pose estimation maps and achieved state-of-the-art results on UTD-MHAD. These methods required that the ``natural'' connectivity between joints was chosen \textit{a priori}. In the proposed method, biomechanical connectivity is not strictly encoded, enabling the exploration of complex spatial relationships between joints, and similarly, temporal relationships between frames. Furthermore, a more seamless integration of pose information is desirable to enable end-to-end training for the task for action recognition. 

To this end, Choutas \textit{et al.}~\cite{choutas2018potion} integrated pose information more coherently using the confidence maps generated by a pose estimation network as the input to a 2D CNN classifier. Pose data returned from such networks is generally 4D, having the shape (frame, height, width, keypoint). Thus, Choutas \textit{et al.} manually reduced the dimensionality of the pose data by summing the individual keypoint activations temporally and encoding with color, such that the resulting clip-level representation could be processed using a 2D CNN. McNally \textit{et al.}~\cite{mcnally2018cvis} explored a different clip-level encoding technique that involved summing the keypoints spatially and compressing the temporal dimension. Although these architectures integrated pose information in a more seamless manner, end-to-end training was never explored in these works. Finally, Luvizon \textit{et al.}~\cite{luvizon20182d} demonstrated that the tasks of human action recognition and pose estimation could be jointly-learned in an end-to-end multitask framework.


Most similar to our model are those of Choutas \textit{et al.} and McNally \textit{et al.} The proposed STAR framework differs from these approaches in three key aspects encompassing the main contributions of this work: 
\begin{itemize}
\item We introduce the idea of leveraging 3D convolutions to reproject spatio-temporal activations generated by pose estimation layers within a network in a direct manner.  This reprojection strategy avoids potential losses of information associated with a manual dimensionality reduction. 
\item The STAR framework integrates a stack of pose estimation layers within a seamless network architecture, which enables end-to-end training. We further perform pose estimation layer ablation experiments in this work.
\item We use top-down pose estimation and a sliding window, which permits simultaneous localization and detection of multiple actions from multiple subjects in a single video clip. 
\end{itemize}

\section{Method}

\begin{figure*}
    \centering
    \includegraphics[width=1\linewidth]{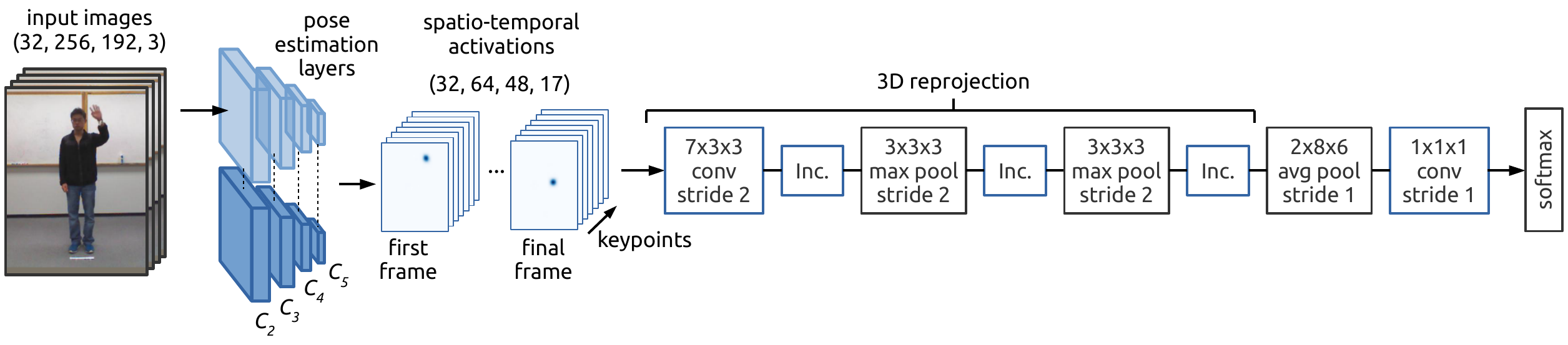}
    \vspace{-20pt}
    \caption{STAR-Net. Within an end-to-end network architecture, a sequence of RGB input images are fed through a stack of pose estimation layers, resulting in a 4D set of spatio-temporal activations. The spatio-temporal activations are then reprojected via a stack of inflated inception~\cite{carreira2017quo} convolutional layers.  Finally, predictions of human action are obtained via a final stack consisting of an average pooling layer, a point-wise convolutional layer, and a softmax layer.}
    \label{fig:arch}
\end{figure*}

In this section we detail the proposed STAR framework. The components of the framework are described as follows. Using a sliding window approach, a person detector is used to locate the subject. The detections are cropped, padded, and resized in accordance with the input size of 256$\times$192. Within an end-to-end network architecture (STAR-Net), a sequence of RGB input images are fed through a stack of pose estimation layers, resulting in a 4D set of spatio-temporal activations. The spatio-temporal activations are then reprojected via a stack of inflated inception convolutional layers~\cite{carreira2017quo}. Finally, predictions of human actions are obtained via a final stack consisting of an average pooling layer, a point-wise convolutional layer, and a softmax layer. The STAR-Net architecture is illustrated in Fig.~\ref{fig:arch}.

\subsection{Person Detection}
As with all top-down pose estimation models, a person detector is generally required. It was found that a simple histogram of oriented gradients (HOG) person detector missed many detections on the J-HMDB dataset. With concerns that missed detections may be detrimental to recognition performance, a more robust deep learning-based detector was chosen, namely the MobileNetV2 Single-shot Multibox Detector~\cite{sandler2018mobilenetv2} (SSD). Using the bounding boxes returned by the SSD, the images were cropped, padded to an aspect ratio of 4:3, and then resized in accordance with the 256$\times$192 input size of the first pose estimation layer using bilinear interpolation. Even using the SSD, detections were occasionally missed. In the case of a missed detection, the bounding box from the previous frame was used. 

Compared to the bottom-up clip-level representation used by Choutas \textit{et al.}~\cite{choutas2018potion}, our top-down sliding window approach enables the simultaneous detection and localization of multiple actions in a single video sample. This is an important feature for many applications, particularly for sports, where it is beneficial to identify when and where actions occur.

\subsection{Pose Estimation Layers}

The pose estimation layers used in the STAR-Net architecture are derived from the Cascaded Pyramid Network~\cite{chen2017cascaded} (CPN) architecture, which placed first in the 2017 and 2018 Microsoft COCO Keypoints Challenge. We chose the CPN~\footnote{TensorFlow CPN implementation available at \url{https://github.com/chenyilun95/tf-cpn}} more for its efficient network architecture than its high precision. In fact, in Section~\ref{sec:starnet_results} we empirically show that high-precision keypoint localization is not a prerequisite for accurate action classification.  

Prior to the release of the CPN, hourglass networks were prevalent for human pose estimation~\cite{newell2016stacked}. The principle of the hourglass architecture lies in repeated bottom-up, top-down processing to consolidate features across multiple scales and encode the local and global context required for the spatial relationships of the human body. These hourglass modules were then stacked with intermediate supervision to improve performance. However, there are computational inefficiencies associated with hourglass stacking as performance gains drop after two stages, leading to wasteful computations in subsequent stages~\cite{chen2017cascaded}. The CPN was designed to mitigate these inefficiencies using a feature pyramid network~\cite{lin2017feature} with a ResNet-50 \cite{he2016deep} backbone. The feature pyramid, which the authors refer to as GlobalNet, makes liberal use of 1$\times$1 convolutions and intermediate supervision. The CPN also uses an adjacent network called RefineNet to refine keypoint predictions. The RefineNet efficiently combines features across multiple scales and as a result, the CPN outperforms an 8-stage hourglass network at less than a third of the computational cost~\cite{chen2017cascaded}. In this paper, we refer to the output feature maps of the feature pyramid as $C_2$ through $C_5$ (see Fig. \ref{fig:arch}).  

In Section~\ref{sec:starnet_results}, a pose layer ablation study is performed to assess the trade-off between action recognition performance and the use of activations from various depths of the CPN. The computational efficiency of the action recognition model can be improved if the RefineNet or feature pyramid blocks can be removed without sacrificing classification accuracy.

\subsection{Spatio-Temporal Activation Reprojection Layers}

The set of 4D spatio-temporal activations (frame, height, width, keypoint) are reprojected in space and time via a stack of 3D convolutional layers.  In essence, the 4D spatio-temporal activations capture the pixel-wise confidence values for the presence of keypoints over time and space. The pose estimation layers were trained to detect the eyes, ears, nose, shoulders, elbows, wrists, hips, knees and ankles (\textit{i.e.}, 17 keypoints in total).  The 3D spatio-temporal activation reprojection subarchitecture shown in Fig.~\ref{fig:arch} begins with a 7$\times$3$\times$3 convolution with a stride of 2, where the filter dimensions are in the format frame$\times$height$\times$width. Following the initial convolution is a series of inflated inception modules~\cite{carreira2017quo} and max pooling layers. 

The inflated inception modules used in the 3D reprojection subarchitecture were based on those introduced in I3D~\cite{carreira2017quo} and follow a similar structure as the modules in the original Inception network~\cite{szegedy2015going}, except with filters inflated to three dimensions. One of the principles behind the inception module is to judiciously reduce dimensionality wherever the computational requirements would increase too much otherwise~\cite{szegedy2015going}. In 2D inception modules, this is accomplished using 1$\times$1 convolutions, whereas in the 3D inception modules it is accomplished using 1$\times$1$\times$1 convolutions. Inception modules reduce the total number of parameters and permit deeper networks to be trained more effectively with less data. Therefore, inception modules help achieve better performance on the small-scale datasets evaluated in Section~\ref{sec:exp}.

\begin{figure*}[t]
    \vspace{-10pt}
    \centering
    \includegraphics[width=0.9\linewidth]{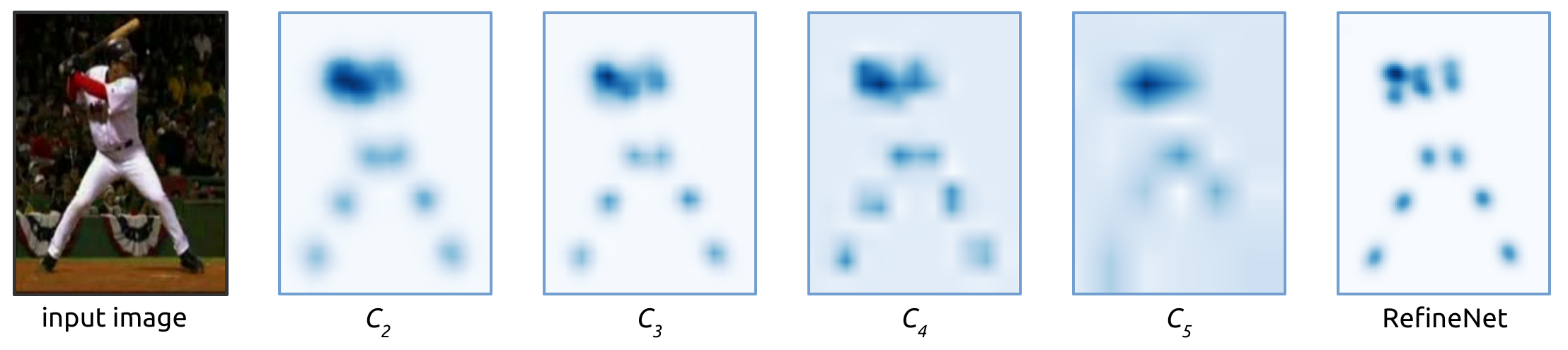}
    \vspace{-10pt}
    \caption{Activations at various depths of the CPN architecture, including the outputs of the feature pyramid blocks $C_2$ to $C_5$, and the final RefineNet output. For display purposes, the eyes and ears were excluded and the activations were summed along the keypoint axis.}
    \label{fig:pose_levels}
    \vspace{-15pt}
\end{figure*}

\subsection{Action Recognition Prediction Layers}

Following the 3D spatio-temporal activation reprojection layers, an average pooling layer with a window of 2$\times$8$\times$6 and stride of 1 is applied, which collapses the spatial dimension. A final 1$\times$1$\times$1 convolution is applied with the output channels equal to the number of action classes. The final size of the temporal dimension will vary based on the number of input frames. Using a 32-frame input results in a temporal feature size of 3. The temporal features are averaged to reach a final prediction, which provides flexibility for the temporal breadth of the input. This characteristic is critical during testing, when long-duration video samples with varying number of frames must be classified using a single label. 

\subsection{Implementation Details}
Each convolutional layer is followed by batch normalization~\cite{ioffe2015batch} and the ReLU non-linearity. The pose estimation layers were initialized with weights pretrained on MSCOCO~\cite{lin2014microsoft}, and the weights of the reprojection and prediction layers were initialized using Gaussian noise. During training, dropout~\cite{srivastava2014dropout} was applied at a rate of 50\% following the average pooling layer. Because the pose estimation subarchitecture accepts batches of frames rather than batches of videos, the spatio-temporal activations were generated prior to training to facilitate batch training of the reprojection layers. We maximize the computational resources of a single NVIDIA Titan Xp GPU using a window size of 32 frames and a batch size of 32. As in~\cite{carreira2017quo}, the start frame was randomly selected during training and, if the number of frames in the video sample was less than 32, the video was looped as many times as necessary. Randomly selecting the start frame served as a form of data augmentation. Other forms of data augmentation included random rotations and horizontal flipping. When flipping the spatial activations, the left and right indices were also switched. For a fair comparison with the similar method of Choutas \textit{et al.}~\cite{choutas2018potion}, only horizontal flipping was used to produce the results on J-HMDB. During testing, predictions are made on the full-length video samples. 

For training, we use the Adam~\cite{kingma2014adam} optimizer and a constant learning rate of 0.001. On a small dataset like UTD-MHAD, the reprojection and prediction layers of STAR-Net can be trained effectively from scratch in just 1000 iterations, and training takes just a few minutes. This is in stark contrast to most state-of-the-art approaches that often require extensive pretraining and large datasets, which can take days of training on multiple GPUs. For example, I3D was trained on Kinetics-400 using 32 GPUs~\cite{carreira2017quo}, although the GPU hours were not reported. 

\section{Experiments}
\label{sec:exp}

In this section we demonstrate the efficacy of STAR-Net for small-scale and single-environment applications through evaluation on UTD-MHAD~\cite{chen2015utd} and J-HMDB~\cite{jhuang2013towards}, two small-scale action recognition datasets containing 27 and 21 action classes, respectively. We compare the results of STAR-Net with other state-of-the-art methods using RGB and other data modalities. 


\subsection{Datasets}

The action classes in \textbf{UTD-MHAD} comprise high-level body segment movements with no scene interaction (\textit{e.g.}, wave, squat, lunge, \textit{etc}); thus UTD-MHAD is a suitable dataset for pose-based action recognition. The actions were performed by 8 subjects, four male and four female, with each subject performing an action four times. After removing three corrupted samples, the dataset includes a total of 861 action samples. We evaluate according to the frequently-used cross-subject protocol, where subjects 1, 3, 5, 7 are used for training and subjects 2, 4, 6, 8 are used for testing. 

\textbf{J-HMDB} is a 21-class subset of the larger HMDB~\cite{kuehne2011hmdb} dataset. Its action classes consist primarily of body segment movements (\textit{e.g.}, golf swing, baseball swing, pull-up, \textit{etc}.) where, for the most part, the full body is visible. For these reasons, J-HMDB is a suitable dataset for pose-based action recognition. J-HMDB contains 928 samples and uses three train/test splits. The results reported on J-HMDB are averaged over the three splits. 

\subsection{STAR-Net Results}
\label{sec:starnet_results}

In this section, we evaluate STAR-Net on the aforementioned datasets and investigate the effects of pose layer ablation and end-to-end training. 

\smallskip\noindent\textbf{Pose Layer Ablation.} For the pose layer ablation, multiple STAR-Net classifiers were trained using pre-generated activations from various levels of the CPN architecture. Specifically, the classifiers were trained using activations from the feature pyramid blocks $C_2$ through $C_5$, and the final output of the RefineNet. The deeper activations with smaller spatial resolutions were upsampled to 64$\times$48.

Fig.~\ref{fig:pose_levels} illustrates the appearance of the activations at different depths of the CPN architecture. Interestingly, the $C_2$ and $C_3$ activations are qualitatively similar to the final output activations of the RefineNet. Motivated by this visual observation, we hypothesize that activations at shallower depths may be leveraged within STAR-Net to enable greater computational efficiency without sacrificing classification accuracy. 


The performances of STAR-Net variants trained on activations from various depths of the CPN are reported in Table~\ref{tab:starnet_results}. On UTD-MHAD, the performance using the RefineNet activations was matched using activations from the feature pyramid blocks $C_2$ and $C_3$. Furthermore, classification accuracy decreased less than 1\% using the relatively high-level pose information encoded in $C_4$ and $C_5$. This result suggests that high-precision pose estimation is not essential within the STAR framework for the purpose of action recognition. This is in contrast to other pose-based approaches that saw significant performance improvements using ground-truth keypoint locations~\cite{choutas2018potion, cheronICCV15, zolfaghari2017chained}. The inference speeds of each end-to-end architecture are also reported in Table~\ref{tab:starnet_results}. As expected, removing pose estimation layers significantly improves computational efficiency. By removing the RefineNet, STAR-Net becomes 28\% faster (STAR-Net-$C_2$ versus STAR-Net-RN). By removing feature pyramid blocks, STAR-Net becomes 42\% faster with minimal loss in performance (STAR-Net-$C_5$ versus STAR-Net-RN). 

\smallskip\noindent\textbf{End-to-end Training.} Considering the significant improvement in forward propagation time of STAR-Net-$C_2$ compared to STAR-Net-RN, and the apparent negligible loss in classification accuracy, we chose to investigate the fine-tuning of STAR-Net-$C_2$ in an end-to-end training manner on UTD-MHAD. To accommodate batch training of the 3D reprojection layers, multiple 32-frame video samples were concatenated along the batch dimension and used as input to the pose estimation layers. Upon reaching the 3D reprojection, the spatio-temporal activations were reshaped to 5D, effectively separating the video samples. Due to limited GPU memory, it was only possible to batch two 32-frame video samples at a time. For this fine-tuning, the input images were additionally augmented using random horizontal and vertical scaling in the range of 0.75 to 1.25 with a probability of 50\%. The intent of the scaling was to account for inter-subject variation in body types. Random scaling was not practical when the spatio-temporal activations were pre-generated, as it would have altered the spatial footprint of the activations, which should remain the same regardless of the scale of the input image.

\begin{figure}
    \centering
    \includegraphics[width=0.8\linewidth]{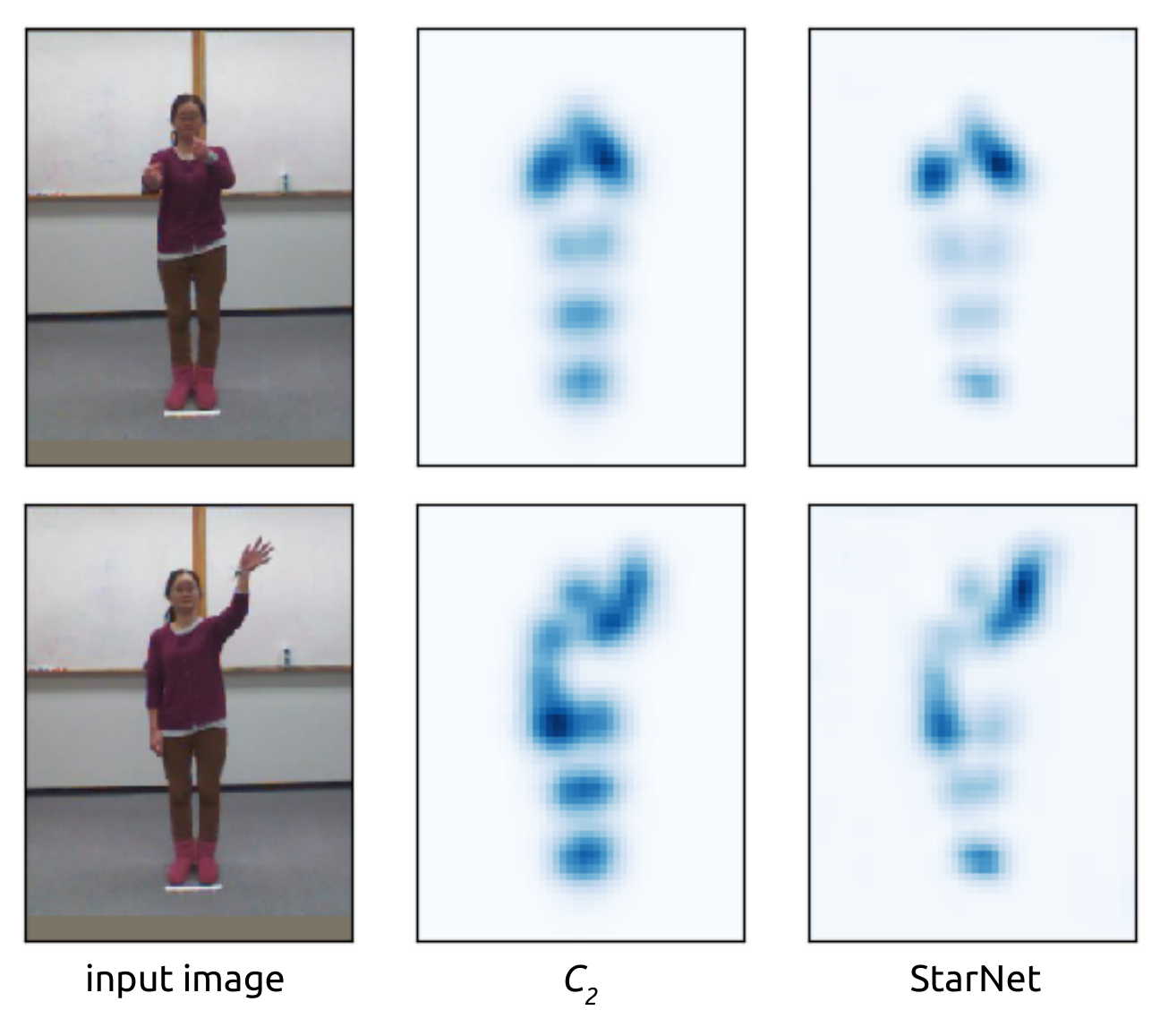}
    \vspace{-15pt}
    \caption{Spatial activations before and after end-to-end fine-tuning of STAR-Net for action recognition. A very interesting observation that can be made is that the keypoints contributing most to the action, such as motions of the arms and feet, are greatly accentuated in the spatial activations of the fine-tuned STAR-Net.}
    \label{fig:conv2_vs_starnet}
    \vspace{-5pt}
\end{figure}

The fine-tuned STAR-Net-$C_2$, which we will refer to as simply STAR-Net moving forward, managed a marginal improvement in classification accuracy on UTD-MHAD, reaching 90.0\%. It is suspected that greater improvements may be achievable with additional GPU memory resources to accommodate a larger video sample batch size. Still, a particular phenomenon worth discussing is the change in the spatial activations generated by the pose estimation layers of STAR-Net following end-to-end training for action recognition. Fig.~\ref{fig:conv2_vs_starnet} illustrates that the spatial activations of keypoints that contributed most to the action, such as motions of the arms and feet were greatly accentuated in the fine-tuned STAR-Net. Conversely, the non-moving keypoints such as the knees and hips were attenuated. This interesting phenomenon can potentially be leveraged to gain new insights into discovering what the key defining characteristics are for particular actions.

\begin{table}
\caption{Classification accuracy of STAR-Net using activations from different pose estimation depths.}
\vspace{-10pt}
\label{tab:starnet_results}
\begin{center}
\begin{tabular}{|l|c|c|c|}
\hline
Method & UTD-MHAD & J-HMDB & ms/sample$^*$\\
\hline\hline
STAR-Net-RN & 88.8 & \textbf{64.3} & 84.0\\
STAR-Net-$C_2$ & 88.8 & 64.0 & 60.4\\
STAR-Net-$C_3$ & 88.8 & 61.5 & 52.9\\
STAR-Net-$C_4$ & 87.4 & 61.6 & 49.7\\
STAR-Net-$C_5$ & 87.9 & 61.6 & \textbf{48.9}\\
\hline
STAR-Net$^\dagger$ & \textbf{90.0} & -- & 60.4\\
\hline
\end{tabular}
\end{center}
$^*$Forward propagation of 32-frame video sample, excluding detection, I/O, and preprocessing. Averaged over 1000 trials using videos from UTD-MHAD. 
$^\dagger$End-to-end fine-tuned STAR-Net using the STAR-Net-$C_2$ architecture.
\vspace{-15pt}
\end{table}

\subsection{Comparison with the State-of-the-art}

UTD-MHAD was chosen with the intent of assessing the ability of STAR-Net to be trained with limited data and recognize actions effectively in single-environments. In single-environment applications, there is minimal variance in the overall appearance of the videos, and therefore an action classifier cannot rely on global appearance cues to make predictions. State-of-the-art RGB-based action recognition models such as I3D~\cite{carreira2017quo} are typically evaluated on large-scale datasets like UCF-101~\cite{soomro2012ucf101} and Kinetics-400~\cite{kay2017kinetics}. Generally speaking, the action classes in these datasets have unique environments (\textit{e.g.}, sky diving, soccer, skiing). Hence, we suspect that models trained on these datasets benefit greatly from global appearance cues. As such, the effectiveness of action recognition models trained on large-scale datasets for single-environment applications remains an open research question. To this end, we elected to fine-tune I3D~\cite{carreira2017quo} on UTD-MHAD. The same detections were used, except we cropped and resized the detections to 224$\times$224 in accordance with the input size of I3D.

The fine-tuned I3D results on UTD-MHAD are reported in Table~\ref{tab:utd_soa} along with other published methods. STAR-Net outperforms the fine-tuned I3D model by 12.8\%. The significant performance gap between the RGB and optical flow streams of I3D is an indication that global appearance is less effective in the single environment. On UCF-101 and Kinetics-400, the performance gap between I3D's two streams is much less (1.1\% and 2.3\%, respectively, in favour of optical flow)~\cite{carreira2017quo}. This insight supports the hypothesis that RGB action recognition networks developed for large-scale datasets rely heavily on global appearance cues, thus causing them to underperform in single-environments. However, the authors acknowledge that the network capacity of I3D is likely too great for UTD-MHAD, and that overfitting may have been a factor contributing to the poor results. 

\begin{table}
\caption{Results on UTD-MHAD cross-subject experiment}
\vspace{-10pt}
\begin{center}
\begin{tabular}{|l|c|}
\hline
Method & Accuracy (\%) \\
\hline\hline
I3D (RGB) & 61.9\\
I3D (Flow) & 71.9\\
I3D (RGB+Flow) & 77.2\\
\hline 
McNally \textit{et al.}~\cite{mcnally2018cvis} (RGB) & 76.1\\
Chen \textit{et al.}~\cite{chen2016fusion} (Depth+Inertial) & 79.1\\
Hussein \textit{et al.}~\cite{hussein2013human} (3D Skeletal) & 85.6\\
Wang \textit{et al.}~\cite{wang2016action} (3D Skeletal) & 85.8\\
Hou \textit{et al.}~\cite{hou2018skeleton} (3D Skeletal) & 87.0\\
Liu \textit{et al.}~\cite{liu2018recognizing} (RGB) & \textbf{92.8}\\
\hline 
STAR-Net & \textbf{90.0}\\
STAR-Net+I3D RGB & 83.7\\
STAR-Net+I3D Flow & 88.1\\
STAR-Net+I3D RGB+Flow & 88.4\\
\hline
\end{tabular}
\end{center}
\label{tab:utd_soa}
\vspace{-20pt}
\end{table}

In \cite{choutas2018potion}, it was demonstrated that pose-based features were complementary to those produced by I3D. To this end, the bottom rows of Table~\ref{tab:utd_soa} show the result of combining the STAR-Net predictions with the two streams of I3D using equal weights. On UTD-MHAD, the effect of combining the models decreased performance. The results in Table~\ref{tab:utd_soa} also show that STAR-Net outperforms several methods using richer data modalities, but surrenders 2.8\% to the state-of-the-art pose-based method of Liu \textit{et al.}~\cite{liu2018recognizing}, who used spatial rank pooling to encode the evolution of 2D pose images and averaged pose heatmaps (\textit{i.e.}, as separate streams). Interestingly, the performance of each stream alone was 85.6\% and 74.9\%, respectively, indicating that these streams were highly complementary. Provided that STAR-Net was able to achieve 89.8\% using only pose heatmaps (before fine-tuning), it is possible that incorporating 2D pose images as a separate stream could be advantageous. This is a proposed area for future research.


In Table~\ref{tab:jhmdb_soa}, STAR-Net is compared to state-of-the-art methods on J-HMDB. The method of Choutas \textit{et al.} (PoTion)~\cite{choutas2018potion} is the most similar to STAR-Net, in that is uses a manual dimensionality reduction and 2D convolutions in place of 3D convolutions. Notably, the use of 3D convolutions has lead to significant performance gains over the former. STAR-Net also outperforms popular action recognition models such as P-CNN~\cite{cheronICCV15} and Action Tubes~\cite{gkioxari2015finding}, but falls short to Peng \textit{et al.}~\cite{peng2016multi} and the state-of-the-art method of Zolfaghari \textit{et al.}~\cite{zolfaghari2017chained}, who used a Markov chain model to sequentially refine predictions from three 3D CNN streams encoding pose, optical flow, and RGB. It is worth noting that in the model of Zolfaghari \textit{et al.}, the pose stream on its own yields a classification accuracy of 45.5\%, which is much lower than STAR-Net (64.3\%). This suggests that the integration of optical flow and RGB streams via the Markov chain was highly effective. STAR-Net may benefit from a similar framework, and thus incorporating optical flow and RGB streams via a Markov chain is another targeted area for future research.

\begin{table}
\caption{Results on J-HMDB (averaged over 3 splits)}
\vspace{-10pt}
\begin{center}
\begin{tabular}{|l|c|}
\hline
Method & Accuracy (\%) \\
\hline\hline
Choutas \textit{et al.} (PoTion)~\cite{choutas2018potion} & 57.0\\
Chéron \textit{et al.} (P-CNN)~\cite{cheronICCV15} & 61.1\\
Gkioxari \textit{et al.} (Action Tubes)~\cite{gkioxari2015finding} & 62.5\\
Peng \textit{et al.} (MR TS R-CNN)~\cite{peng2016multi} & 71.1\\
Zolfaghari \textit{et al.} (Chained)~\cite{zolfaghari2017chained} & \textbf{76.1}\\
\hline
STAR-Net & 64.3\\
\hline 
\end{tabular}
\end{center}
\label{tab:jhmdb_soa}
\vspace{-20pt}
\end{table}

\section{Conclusion}
This paper introduces the concept of spatio-temporal activation reprojection for human action recognition. The STAR framework seamlessly integrates human pose estimation with 3D convolutional reprojection in a coherent end-to-end architecture to efficiently detect actions in video. Empirical results on multiple action recognition datasets demonstrated that precise pose estimation is not essential within the STAR framework. This facilitated the compression of the pose estimation subarchitecture, leading to improved inference speed. Through evaluation on a multimodal action recognition dataset, it was shown that our approach is superior to many action recognition methods utilizing multiple richer data modalities, including depth and inertial sensors. Using a sliding window approach with top-down pose estimation, our method permits the simultaneous detection and localization of actions in continuous video streams.

\section*{Acknowledgment}
We would like to acknowledge financial support from the Canada Research Chairs program and the Natural Sciences and Engineering Research Council of Canada (NSERC). We would also like to acknowledge the NVIDIA GPU Grant Program for the Titan Xp GPU donation.



\bibliographystyle{IEEEtran}
%

\bibliography{starnet}

\begin{thebibliography}{10}
\providecommand{\url}[1]{#1}
\csname url@samestyle\endcsname
\providecommand{\newblock}{\relax}
\providecommand{\bibinfo}[2]{#2}
\providecommand{\BIBentrySTDinterwordspacing}{\spaceskip=0pt\relax}
\providecommand{\BIBentryALTinterwordstretchfactor}{4}
\providecommand{\BIBentryALTinterwordspacing}{\spaceskip=\fontdimen2\font plus
\BIBentryALTinterwordstretchfactor\fontdimen3\font minus
  \fontdimen4\font\relax}
\providecommand{\BIBforeignlanguage}[2]{{%
\expandafter\ifx\csname l@#1\endcsname\relax
\typeout{** WARNING: IEEEtran.bst: No hyphenation pattern has been}%
\typeout{** loaded for the language `#1'. Using the pattern for}%
\typeout{** the default language instead.}%
\else
\language=\csname l@#1\endcsname
\fi
#2}}
\providecommand{\BIBdecl}{\relax}
\BIBdecl

\bibitem{ji20133d}
S.~Ji, W.~Xu, M.~Yang, and K.~Yu, ``3d convolutional neural networks for human
  action recognition,'' \emph{IEEE transactions on pattern analysis and machine
  intelligence}, 2013.

\bibitem{fani2017hockey}
M.~Fani, H.~Neher, D.~A. Clausi, A.~Wong, and J.~Zelek, ``Hockey action
  recognition via integrated stacked hourglass network,'' in \emph{CVPR
  Workshops}, 2017.

\bibitem{mitra2007gesture}
S.~Mitra and T.~Acharya, ``Gesture recognition: A survey,'' \emph{IEEE
  Transactions on Systems, Man, and Cybernetics, Part C (Applications and
  Reviews)}, 2007.

\bibitem{chen2016fusion}
C.~Chen, R.~Jafari, and N.~Kehtarnavaz, ``Fusion of depth, skeleton, and
  inertial data for human action recognition,'' in \emph{ICASSP}, 2016.

\bibitem{du2015hierarchical}
Y.~Du, W.~Wang, and L.~Wang, ``Hierarchical recurrent neural network for
  skeleton based action recognition,'' in \emph{CVPR}, 2015.

\bibitem{liu2016spatio}
J.~Liu, A.~Shahroudy, D.~Xu, and G.~Wang, ``Spatio-temporal lstm with trust
  gates for 3d human action recognition,'' in \emph{ECCV}, 2016.

\bibitem{mehta2017vnect}
D.~Mehta, S.~Sridhar, O.~Sotnychenko, H.~Rhodin, M.~Shafiei, H.-P. Seidel,
  W.~Xu, D.~Casas, and C.~Theobalt, ``Vnect: Real-time 3d human pose estimation
  with a single rgb camera,'' \emph{ACM Transactions on Graphics}, 2017.

\bibitem{krizhevsky2012imagenet}
A.~Krizhevsky, I.~Sutskever, and G.~E. Hinton, ``Imagenet classification with
  deep convolutional neural networks,'' in \emph{NIPS}, 2012.

\bibitem{donahue2015long}
J.~Donahue, L.~Anne~Hendricks, S.~Guadarrama, M.~Rohrbach, S.~Venugopalan,
  K.~Saenko, and T.~Darrell, ``Long-term recurrent convolutional networks for
  visual recognition and description,'' in \emph{CVPR}, 2015.

\bibitem{peng2016multi}
X.~Peng and C.~Schmid, ``Multi-region two-stream r-cnn for action detection,''
  in \emph{ECCV}, 2016.

\bibitem{karpathy2014large}
A.~Karpathy, G.~Toderici, S.~Shetty, T.~Leung, R.~Sukthankar, and L.~Fei-Fei,
  ``Large-scale video classification with convolutional neural networks,'' in
  \emph{CVPR}, 2014.

\bibitem{newell2016stacked}
A.~Newell, K.~Yang, and J.~Deng, ``Stacked hourglass networks for human pose
  estimation,'' in \emph{ECCV}, 2016.

\bibitem{cao2017realtime}
Z.~Cao, T.~Simon, S.-E. Wei, and Y.~Sheikh, ``Realtime multi-person 2d pose
  estimation using part affinity fields,'' in \emph{CVPR}, 2017.

\bibitem{chen2017cascaded}
Y.~Chen, Z.~Wang, Y.~Peng, Z.~Zhang, G.~Yu, and J.~Sun, ``Cascaded pyramid
  network for multi-person pose estimation,'' in \emph{CVPR}, 2018.

\bibitem{chen2015utd}
C.~Chen, R.~Jafari, and N.~Kehtarnavaz, ``Utd-mhad: A multimodal dataset for
  human action recognition utilizing a depth camera and a wearable inertial
  sensor,'' in \emph{ICIP}, 2015.

\bibitem{jhuang2013towards}
H.~Jhuang, J.~Gall, S.~Zuffi, C.~Schmid, and M.~J. Black, ``Towards
  understanding action recognition,'' in \emph{ICCV}, 2013.

\bibitem{blank2005actions}
M.~Blank, L.~Gorelick, E.~Shechtman, M.~Irani, and R.~Basri, ``Actions as
  space-time shapes,'' in \emph{ICCV}, 2005.

\bibitem{schuldt2004recognizing}
C.~Schuldt, I.~Laptev, and B.~Caputo, ``Recognizing human actions: a local svm
  approach,'' in \emph{ICPR}, 2004.

\bibitem{scovanner20073}
P.~Scovanner, S.~Ali, and M.~Shah, ``A 3-dimensional sift descriptor and its
  application to action recognition,'' in \emph{ACM international conference on
  Multimedia}, 2007.

\bibitem{klaser2008spatio}
A.~Klaser, M.~Marsza{\l}ek, and C.~Schmid, ``A spatio-temporal descriptor based
  on 3d-gradients,'' in \emph{BMVC}, 2008.

\bibitem{willems2008efficient}
G.~Willems, T.~Tuytelaars, and L.~Van~Gool, ``An efficient dense and
  scale-invariant spatio-temporal interest point detector,'' in \emph{ECCV},
  2008.

\bibitem{yeffet2009local}
L.~Yeffet and L.~Wolf, ``Local trinary patterns for human action recognition,''
  in \emph{ICCV}, 2009.

\bibitem{wang2013action}
H.~Wang and C.~Schmid, ``Action recognition with improved trajectories,'' in
  \emph{CVPR}, 2013.

\bibitem{tran2015learning}
D.~Tran, L.~Bourdev, R.~Fergus, L.~Torresani, and M.~Paluri, ``Learning
  spatiotemporal features with 3d convolutional networks,'' in \emph{ICCV},
  2015.

\bibitem{carreira2017quo}
J.~Carreira and A.~Zisserman, ``Quo vadis, action recognition? a new model and
  the kinetics dataset,'' in \emph{CVPR}, 2017.

\bibitem{wang2012mining}
J.~Wang, Z.~Liu, Y.~Wu, and J.~Yuan, ``Mining actionlet ensemble for action
  recognition with depth cameras,'' in \emph{CVPR}, 2012.

\bibitem{hussein2013human}
M.~E. Hussein, M.~Torki, M.~A. Gowayyed, and M.~El-Saban, ``Human action
  recognition using a temporal hierarchy of covariance descriptors on 3d joint
  locations.'' in \emph{IJCAI}, 2013.

\bibitem{xia2012view}
L.~Xia, C.-C. Chen, and J.~K. Aggarwal, ``View invariant human action
  recognition using histograms of 3d joints,'' in \emph{CVPR Workshops}, 2012.

\bibitem{wang2016action}
P.~Wang, Z.~Li, Y.~Hou, and W.~Li, ``Action recognition based on joint
  trajectory maps using convolutional neural networks,'' in \emph{ACM on
  Multimedia Conference}, 2016.

\bibitem{hou2018skeleton}
Y.~Hou, Z.~Li, P.~Wang, and W.~Li, ``Skeleton optical spectra-based action
  recognition using convolutional neural networks,'' \emph{IEEE Transactions on
  Circuits and Systems for Video Technology}, 2018.

\bibitem{yan2018spatial}
S.~Yan, Y.~Xiong, and D.~Lin, ``Spatial temporal graph convolutional networks
  for skeleton-based action recognition,'' in \emph{AAAI}, 2018.

\bibitem{liu2018recognizing}
M.~Liu and J.~Yuan, ``Recognizing human actions as the evolution of pose
  estimation maps,'' in \emph{CVPR}, 2018.

\bibitem{choutas2018potion}
V.~Choutas, P.~Weinzaepfel, J.~Revaud, and C.~Schmid, ``Potion: Pose motion
  representation for action recognition,'' in \emph{CVPR}, 2018.

\bibitem{mcnally2018cvis}
W.~McNally, A.~Wong, and J.~McPhee, ``Action recognition using deep
  convolutional neural networks and compressed spatio-temporal pose
  encodings,'' in \emph{CVIS}, 2018.

\bibitem{luvizon20182d}
D.~C. Luvizon, D.~Picard, and H.~Tabia, ``2d/3d pose estimation and action
  recognition using multitask deep learning,'' in \emph{CVPR}, 2018.

\bibitem{sandler2018mobilenetv2}
M.~Sandler, A.~Howard, M.~Zhu, A.~Zhmoginov, and L.-C. Chen, ``Mobilenetv2:
  Inverted residuals and linear bottlenecks,'' in \emph{CVPR}, 2018.

\bibitem{lin2017feature}
T.-Y. Lin, P.~Doll{\'a}r, R.~Girshick, K.~He, B.~Hariharan, and S.~Belongie,
  ``Feature pyramid networks for object detection,'' in \emph{CVPR}, 2017.

\bibitem{he2016deep}
K.~He, X.~Zhang, S.~Ren, and J.~Sun, ``Deep residual learning for image
  recognition,'' in \emph{CVPR}, 2016.

\bibitem{szegedy2015going}
C.~Szegedy, W.~Liu, Y.~Jia, P.~Sermanet, S.~Reed, D.~Anguelov, D.~Erhan,
  V.~Vanhoucke, and A.~Rabinovich, ``Going deeper with convolutions,'' in
  \emph{CVPR}, 2015.

\bibitem{ioffe2015batch}
S.~Ioffe and C.~Szegedy, ``Batch normalization: Accelerating deep network
  training by reducing internal covariate shift,'' \emph{arXiv preprint
  arXiv:1502.03167}, 2015.

\bibitem{lin2014microsoft}
T.-Y. Lin, M.~Maire, S.~Belongie, J.~Hays, P.~Perona, D.~Ramanan,
  P.~Doll{\'a}r, and C.~L. Zitnick, ``Microsoft coco: Common objects in
  context,'' in \emph{ECCV}, 2014.

\bibitem{srivastava2014dropout}
N.~Srivastava, G.~Hinton, A.~Krizhevsky, I.~Sutskever, and R.~Salakhutdinov,
  ``Dropout: a simple way to prevent neural networks from overfitting,''
  \emph{The Journal of Machine Learning Research}, 2014.

\bibitem{kingma2014adam}
D.~P. Kingma and J.~Ba, ``Adam: A method for stochastic optimization,''
  \emph{arXiv preprint arXiv:1412.6980}, 2014.

\bibitem{kuehne2011hmdb}
H.~Kuehne, H.~Jhuang, E.~Garrote, T.~Poggio, and T.~Serre, ``Hmdb: a large
  video database for human motion recognition,'' in \emph{ICCV}, 2011.

\bibitem{cheronICCV15}
G.~Ch{\'e}ron, I.~Laptev, and C.~Schmid, ``{P-CNN: Pose-based CNN Features for
  Action Recognition},'' in \emph{ICCV}, 2015.

\bibitem{zolfaghari2017chained}
M.~Zolfaghari, G.~L. Oliveira, N.~Sedaghat, and T.~Brox, ``Chained multi-stream
  networks exploiting pose, motion, and appearance for action classification
  and detection,'' in \emph{ICCV}, 2017.

\bibitem{soomro2012ucf101}
K.~Soomro, A.~R. Zamir, and M.~Shah, ``Ucf101: A dataset of 101 human actions
  classes from videos in the wild,'' \emph{arXiv preprint arXiv:1212.0402},
  2012.

\bibitem{kay2017kinetics}
W.~Kay, J.~Carreira, K.~Simonyan, B.~Zhang, C.~Hillier, S.~Vijayanarasimhan,
  F.~Viola, T.~Green, T.~Back, P.~Natsev \emph{et~al.}, ``The kinetics human
  action video dataset,'' \emph{arXiv preprint arXiv:1705.06950}, 2017.

\bibitem{gkioxari2015finding}
G.~Gkioxari and J.~Malik, ``Finding action tubes,'' in \emph{CVPR}, 2015.

\end{thebibliography}

\end{document}